# Batch Size Influence on Performance of Graphic and Tensor Processing Units during Training and Inference Phases

Yuriy Kochura, Yuri Gordienko, Vlad Taran, Nikita Gordienko, Alexandr Rokovyi, Oleg Alienin, Sergii Stirenko

National Technical University of Ukraine "Igor Sikorsky Kyiv Polytechnic Institute", Kyiv, Ukraine
`iuriy.kochura@gmail.com`

**Abstract.** The impact of the maximally possible batch size (for the better runtime) on performance of graphic processing units (GPU) and tensor processing units (TPU) during training and inference phases is investigated. The numerous runs of the selected deep neural network (DNN) were performed on the standard MNIST and Fashion-MNIST datasets. The significant speedup was obtained even for extremely low-scale usage of Google TPUv2 units (8 cores only) in comparison to the quite powerful GPU NVIDIA Tesla K80 card with the speedup up to 10x for training stage (without taking into account the overheads) and speedup up to 2x for prediction stage (with and without taking into account overheads). The precise speedup values depend on the utilization level of TPUv2 units and increase with the increase of the data volume under processing, but for the datasets used in this work (MNIST and Fashion-MNIST with images of sizes 28x28) the speedup was observed for batch sizes >512 images for training phase and >40 000 images for prediction phase. It should be noted that these results were obtained without detriment to the prediction accuracy and loss that were equal for both GPU and TPU runs up to the 3rd significant digit for MNIST dataset, and up to the 2nd significant digit for Fashion-MNIST dataset.

## 1 Introduction

Now the current trend of hardware acceleration for deep learning applications is largely related with graphics processing units (GPU) as general purpose processors (GPGPU). But recently the interest to alternative platforms, like variety of alternative hardware including GPUs, GPPs, field programmable gate-arrays (FPGA), and digital signal processors (DSP), and application-specific integrated circuit (ASIC) architectures like NVIDIA TCU (Tensor Core Units) and Google Cloud TPU (tensor processing units), become more and more popular [1]. Despite availability of some performance tests, like Google TPU vs. NVIDIA GPU K80 [2] and Google Cloud TPUv2 vs. GPU NVIDIA V100 [3], the systematic studies on the scaling their per-formance (accuracy, loss, inference time, etc.) in relation to datasets of different sizes and hyper-parameters (for example, different batch sizes) are absent. This especially important in the view of the great interest to the influence of hyper-parameters of deep neural networks (DNN) on their training runtime and performance [4-5], especially with regard to the batch size, learning rate, activation functions, etc. [6-9].

The main aim of this paper is to investigate scaling of training and inference performance for the available GPUs and TPUs with an increase of batch size and dataset size. The section *2.Background and Related Work* gives the brief outline of the state of the art in tensor processing and equipment used. The section *3.Experimental and Computational Details* contains the description of the experimental part related with the selected datasets, networks, and metrics used. The section *4.Results* reports about the experimental results obtained, the section *5.Discussion* is dedicated to discussion of these results, and section *6.Conclusions* summarizes the lessons learned.

## 2 Background and Related Work

Google's Tensor Processing Unit (TPU) has recently gained attention as a new and novel approach to increasing the efficiency and speed of neural network processing. According to Google, the TPU can compute neural networks up to 30x faster and up to 80x more power efficient than CPU's or GPU's performing similar applications [2]. It is possible, because the TPU is specifically adapted to solve inference problems with the much higher number of instructions per cycle in comparison to CPU, CPIU with advanced vector extensions, and GPU (Table 1).

**Table 1. Comparison of the number of instructions per cycle for CPU, GPU and TPU**

| Type | Instructions per cycle |
|---|---|
| CPU | ~ $10^0$ |
| CPU (with vector extensions) | ~ $10^1$ |
| GPU | ~ $10^4$ |
| TPU | ~ $10^5$ |

The TPU has a systolic array that contains $256 \times 256 =$ total 65,536 arithmetic logic unit (ALUs), and it can process 65,536 multiply-and-add operations for 8-bit integers every cycle. As far as the TPU runs at 700MHz, it can compute $65,536 \times 7 \times 10^8 = 46 \times 10^{12}$ multiply-and-add operations or $92 \times 10^{12}$ per second [2].

One of the ways to achieve the highest performance in GPU computing is to hide the long latency and other computational overheads by high data-level parallelism to achieve a high throughput, for example by the high batch size values [10-11]. In addition to tests on GPU [4-9], recently the thorough performance analysis of the Google TPU was performed with some attempts to estimate influence of hyper-parameters on performance for TPU also [2,12]. In addition to it this work is aimed to give the answer to some questions, namely, when it could be more efficient to use GPU or TPU during training and inference phases for datasets of various sizes and batch sizes. In the next section the short description of the used datasets, network, equipment, and measurement methods is given.

## 3      Experimental and Computational Details

*Datasets.* The MNIST database (Modified National Institute of Standards and Technology database) is a large database of handwritten digits (28x28 images) that become a standard benchmark for learning, classification and computer vision systems [13]. It was derived from a larger dataset known as the NIST Special Database 19 which contains digits, uppercase and lowercase handwritten letters.

Fashion-MNIST, a new dataset comprising of 28x28 grayscale images of 70,000 fashion products from 10 categories, with 7,000 images per category [14]. The training set has 60,000 images and the test set has 10,000 images. Fashion-MNIST is intended to serve as a direct drop-in replacement for the original MNIST dataset for benchmarking machine learning algorithms, as it shares the same image size, data format and the structure of training and testing splits.

The subsets of these datasets were used with the maximally possible batch size (for the better runtime) starting from 8 images and up to 130 000 images (with duplication of some images to increase the range of the datasets).

*Equipment: GPU and TPU.* The GPU and TPU computing resources were used to investigate the influence of hardware-supported quantization on performance of the DNNs. NVIDIA Tesla K80 was used as GPU cards during these experiments as Google Collaborative cloud resources (https://colab.research.google.com). Google TPUv2 are arranged into 4-chip modules with a performance of 180 TFLOPS, and 64 of these modules are then assembled into 256 chip pods with 11.5 PFLOPS of overall performance. TPU 2.0 has an instruction set optimized for executing Tensorflow and capable of both training and running DNNs. A cloud TPUv2 version was used as a TPU-hardware during these experiments, where 8 TPU cores were available as Google Collaborative cloud resources also.

*Metrics.* Accuracy and loss values are calculated for training, validation, and inference phases, then receiver operating characteristic (ROC) curves are constructed and the area under curve (AUC) is calculated per class and as their micro and macro averages (see below in the next section). To emphasize the contribution of initialization phase for GPU and TPU, the following two runtimes (both for GPU and TPU) per image were calculated for each run:

- time with overheads = the wall time of the $1^{st}$ epoch / number of images;
- time without overheads = the wall time of the $2^{nd}$ epoch / number of images.

The speedup values were calculated as GPU runtimes divided by TPU runtimes.

*Deep Neural Network.* The following deep convolutional neural network (Fig. 1) was used for this stage of research. The idea behind it was to use simple DNN to get results for reasonable period, but DNN should be complex enough to get the high accuracy and low loss comparable with available results on these standard datasets by other researchers.

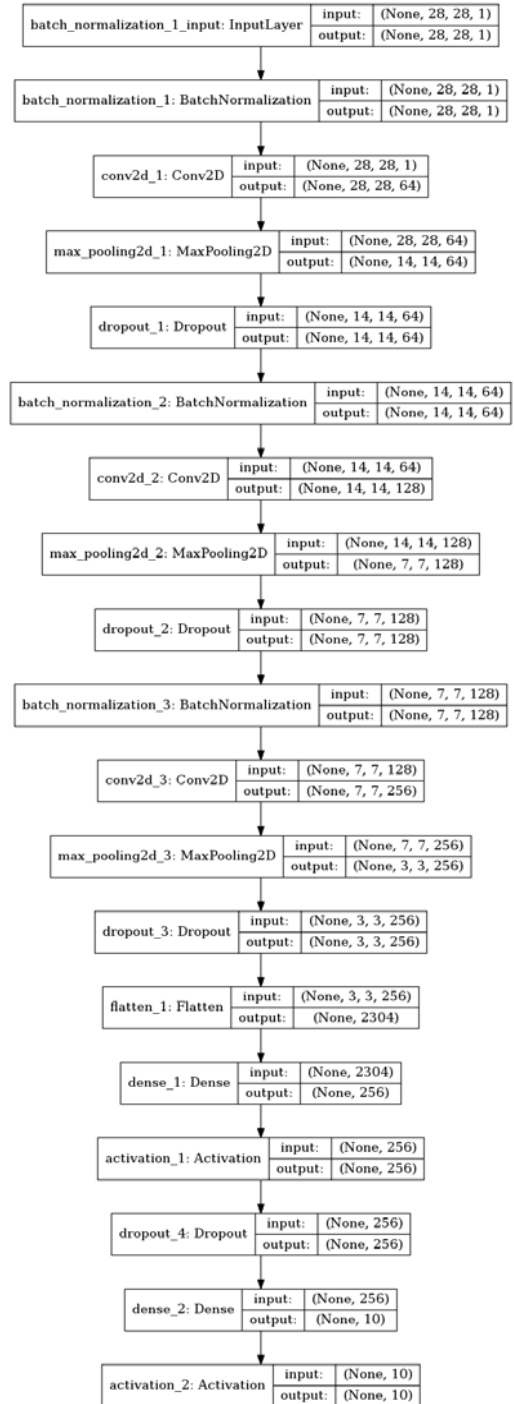

**Fig. 1.** The structure of the deep neural network used in the work.

## 4 Results

### 4.1 GPU

Below the training and validation history is shown for GPU for MNIST dataset (Fig. 2) and the similar plots were obtained for Fashion-MNIST (they are not shown here because of the shortage of space, but will be published elsewhere [12]).

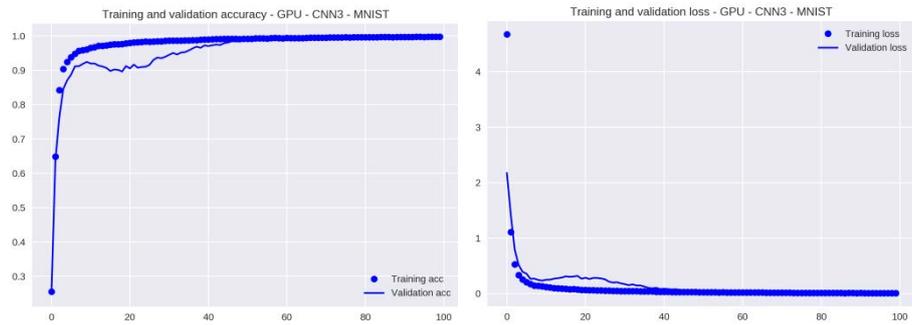

**Fig. 2.** Accuracy (left) and loss (right) during training and validation on GPU K80 (for the whole training part of MNIST dataset 60000 images)

The ROC-curves and AUC-values (Fig. 3) demonstrate the excellent prediction accuracy, which is used for comparison with the similar experiments on TPUv2.

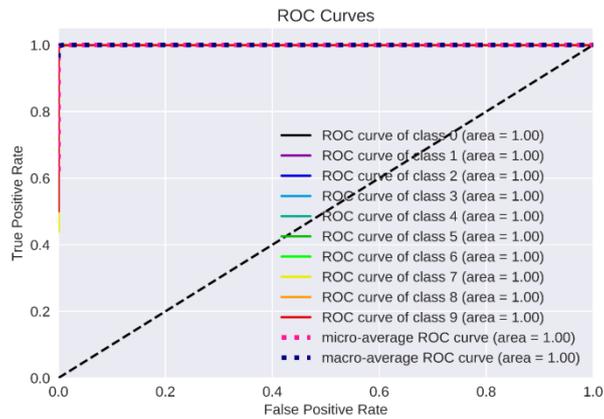

**Fig. 3.** ROC-curves and AUC-values for 10 classes on GPU K80 (for the testing part of MNIST dataset - 10000 images)

### 4.2 TPU

Below the similar training and validation history is shown for TPUv2 and MNIST dataset (Fig. 4) and the similar plots were obtained for Fashion-MNIST (again, they are not shown here because of the shortage of space, but will be published elsewhere [12]). These results demonstrate the principally similar evolution to the high values of accuracy and loss as for GPU before.

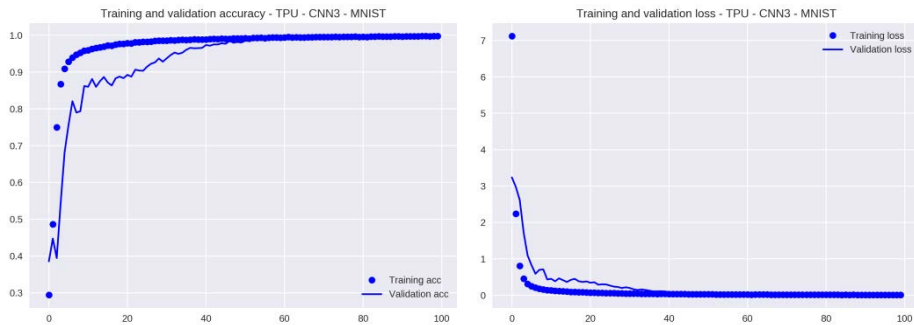

**Fig. 4.** Accuracy (left) and loss (right) during training and validation on Google TPUv2 (for the whole training part of MNIST dataset 60000 images)

The ROC-curves and AUC-values (Fig. 5) also demonstrate the excellent prediction accuracy, which is similar to the results obtained in experiments on GPU.

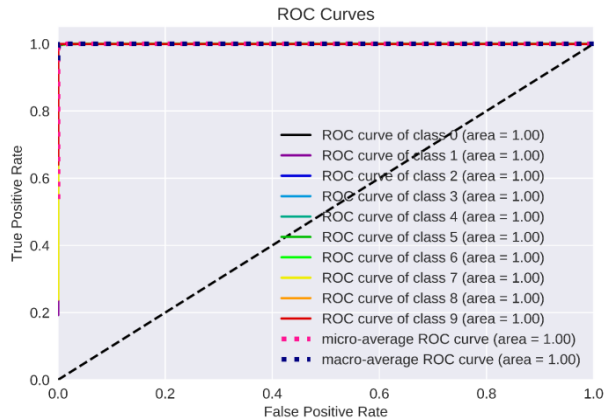

**Fig. 5.** ROC-curves and AUC-values for 10 classes on Google TPUv2 (for the testing part of MNIST dataset - 10000 images)

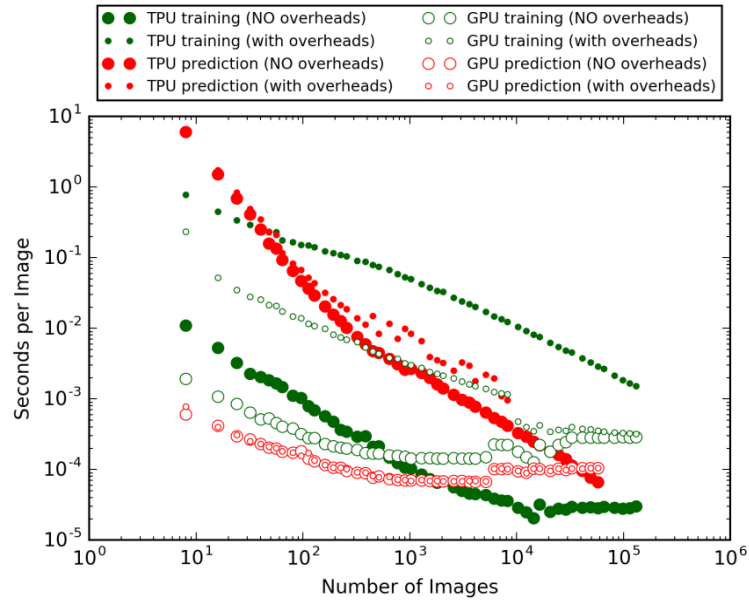

**Fig. 6.** Comparison of GPU K80 and TPUv2 run times per image.

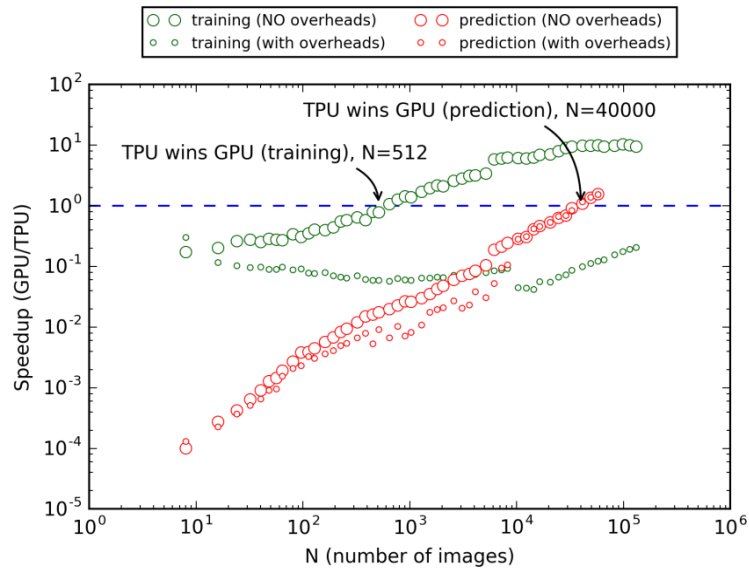

**Fig. 7.** Speedup of GPU K80 and TPUv2.

The saturation of speedup can be observed for training from the raw timing data (Fig. 6) and speedup plot (Fig. 7) in the point where the number of images equal to ~8000. It is explained by the inability to use batch size>8192 for the current datasets with images of sizes 28x28. For the bigger images the maximally possible batch size will be lower and the speedup will be saturated earlier, but it will be investigated elsewhere [15].

## 5  Discussion

These results demonstrate that usage of Google TPUv2 is more effective (faster) than GPU for the large number of computations under conditions of low overhead calculations and high utilization of TPU units. Moreover, these results were obtained for the simple CNN-like deep learning network without detriment to the accuracy and loss that were equal for both GPU and TPU runs up to the 3rd significant digit for MNIST dataset, and up to the 2nd digit for Fashion-MNIST dataset (Table 2).

**Table 2. The accuracy and loss or the testing part (10000 images) of the datasets used.**

| Hardware | Accuracy | Loss |
|---|---|---|
| **MNIST** | | |
| GPU | 0.9944 | 0.02236 |
| TPU | 0.9937 | 0.02214 |
| **Fashion-MNIST** | | |
| GPU | 0.9255 | 0.2354 |
| TPU | 0.9279 | 0.2434 |

The prediction accuracy values were equal for both GPU and TPU up to the 3rd significant digit for MNIST, and up to the 2nd significant digit for Fashion-MNIST. The loss values were equal for both GPU and TPU regimes up to the 2nd significant digit for both MNIST and Fashion-MNIST datasets. The significant speedup (Fig. 7) was reached even for extremely low-scale usage of Google TPUv2 units (8 cores only) in comparison to the quite powerful GPU unit (NVIDIA Tesla K80):
- speedup >10x for training stage (without taking into account overheads);
- speedup >up to 2x for prediction stage (with and without overheads).

The speedup values depend on the utilization level of TPUv2 units and increase with the data volume (batch size) under processing, but for the MNIST and Fashion-MNIST datasets with images of sizes 28x28 the speedup was started (i.e. speedup becomes > 1) after 512 images (for training) and 40 000 images (for prediction) even.

It should be noted that these results were obtained without detriment to the accuracy and loss for the relatively simple DNN and small images (28x28). The current investigations of network size impact and image size impact are under work now and their results will be published elsewhere [15]. In addition to Google TPU architecture, the specific tensor processing hardware tools are available in the other modern GPU-cards like Tesla V100 and Titan V by NVIDIA based on the Volta microarchitecture with specialized Tensor Cores Units (640 TCU) and their influence on training and prediction speedup are under investigation and will be reported elsewhere [15].

As far as the model size limits the available memory space for the batch of images other techniques could be useful for squeezing the model size, like quantization [15-16] and pruning [17-18]. These results can be used for selection of the optimal parameters for applications where a large batch of data needs to be processed, for example, for monitoring real-time road condition in advanced driver assistance systems (ADAS), where such specialized architectures like TPU, TCU, and FPGA-based solutions can be deployed [19]. For example, it is especially important for various complex ADAS-related tasks like real-time obstacle and lane detection [20-21], traffic video analysis under impact of noise [22], and monitoring driver behavior even [23].

# 6    Conclusions

In this work the influence of dataset size with the maximally possible batch size (for the better runtime) was investigated with regard to the performance of graphic and tensor processing hardware during training and inference phases. During these experiments GPU NVIDIA Tesla K80 was used as a GPU-hardware and Google TPUv2 was used as a TPU-hardware (both were available as Google Collaborative cloud resources). The practical efficiency of the both hardware types was investigated during the numerous runs of the selected DNN on the standard MNIST and Fashion-MNIST datasets. The significant speedup was obtained even for extremely low-scale usage of Google TPUv2 units (8 cores only) in comparison to the quite powerful GPU NVIDIA Tesla K80 card with the speedup up to 10x for training stage (without taking into account overheads) and speedup up to 2x for prediction stage (with and without taking into account overheads). It was shown that the precise speedup values depend on the utilization level of TPUv2 units and increase with the increase of the data volume under processing, but for the datasets used in this work (MNIST and Fashion-MNIST with images of sizes 28x28) the speedup was observed for datasets with >512 images for training phase and >40 000 images for prediction phase. In general, the usage of tensor processing architectures like Google TPUv2 will be very promising way for increasing performance of inference and training even, especially in the view of availability of the similar specific tensor processing hardware architectures like TCU in Tesla V100 and Titan V provided by NVIDIA, and others.